\documentclass[11pt]{article}

\usepackage[preprint]{acl}

\usepackage{times}
\usepackage{latexsym}

\usepackage[T1]{fontenc}

\usepackage[utf8]{inputenc}

\usepackage{microtype}

\usepackage{inconsolata}

\usepackage{graphicx}

\usepackage{xcolor}
\usepackage{amsmath}
\usepackage{amssymb}
\usepackage{subcaption}
\usepackage{booktabs}
\usepackage{xurl}
\usepackage{hyperref} 
\usepackage{cleveref}

\title{An Analysis of Residual-Stream Geometry Across Transformer Depth}

\author{
 \textbf{Sunit Bhattacharya},
 \textbf{Ravi Kolli}
\\
 ProRata AI
\\
 \small{
   \textbf{Correspondence:} \href{mailto:sunit@prorata.ai}{sunit@prorata.ai}
 }
}

\begin{document}
\maketitle

\begin{abstract}
We propose a transition-centred geometric analysis of transformer residual
streams.
Relative displacement measures how \emph{far} representations move between
consecutive layers, and orthogonal Procrustes analysis separates each
transition into a rigid rotation and a non-rigid residual.
Across six instruction-tuned models, on code generation and cross-lingual
translation, these measurements reveal reproducible depth regularities.
Relative displacement is strongly layer-dependent; typically larger early
and late, with a quieter middle third; and nearly invariant across
conditions within each model.
Rotation magnitude is nearly constant across depth, while Procrustes residual
and angle concentration remain depth-modulated, with residual peaking at the
final transition.
During generation, non-English targets show larger final-layer displacement
and residual than English targets.
We present these as descriptive geometric regularities, not as measures of
computational effort or causal explanations.
The contribution is a measurement framework for residual-stream transitions
and evidence that, in the settings studied here, depth curves are
model-dependent and largely condition-stable.
\end{abstract}

\section{Introduction}
Transformers~\citep{vaswani2017attention} process information by repeatedly
updating a residual stream \citep{elhage2021mathematical,olah2020overview}.
Most analyses of that stream ask what a layer represents or predicts: representational similarity~\citep{kornblith2019similarity}, vocabulary projections such as the logit lens and tuned lens \citep{nostalgebraist2020logitlens,belrose2023eliciting}, or circuit-level accounts of specific computations \citep{olah2020overview,elhage2021mathematical,bushnaq2024local}.
We propose a complementary view: treat each layer transition as a geometric transformation of the token cloud in residual space.
Under this view, the basic object of study is not a static layer activation,
but the map from the cloud at depth $\ell$ to the cloud at depth $\ell{+}1$.
Relative displacement asks how far tokens move.
Orthogonal Procrustes analysis~\citep{schonemann1966generalized} then
separates that movement into an optimal rigid rotation and a non-rigid
residual.
Recent geometric work on transformers has studied kinematic acceleration
\citep{fernando2025transformer}, curvature along token paths
\citep{damirchi2026truth}, and the intrinsic dimension of per-layer clouds
\citep{viswanathan2025geometry}.
These approaches either snapshot a layer or follow individual tokens.
Our contribution is to measure the collective inter-layer transformation of
the full residual-stream cloud.

We apply this framework to six instruction-tuned models: Qwen3
\(0.6B-8B\), Gemma-2-2B-IT, and StableLM-2-1.6B-Chat, on code generation
and translation between English and four Indo-European languages.
In these settings, four regularities emerge:
\begin{enumerate}
    \item Relative displacement is strongly structured by depth: updates are
    typically larger early and late, with a quieter middle third.
    Within each model, coding and translation conditions share essentially
    the same depth curve.
    \item The magnitude of the global Procrustes rotation is nearly constant
    across depth (variation below a predeclared \(3\%\) bound), while
    Procrustes residual and angle concentration remain depth-modulated, with
    residual peaking at the final transition.
    \item Conditions mainly rescale late-layer amplitude rather than rewrite
    the depth curve.
    During generation, non-English targets show larger final-layer
    displacement and residual than English targets.
    \item These patterns are statistically robust under hierarchical bootstrap
    and layer-permutation tests with Holm correction.
\end{enumerate}
Our contribution is measurement-first.
We introduce a transition-centred geometric view of residual streams and
document regularities that are stable across the models and conditions
studied here.
We do not claim that these metrics measure computational effort, nor do we
offer a causal account of why the patterns arise.
Near-constant rotation magnitude, in particular, may partly reflect
high-dimensional Procrustes geometry; the more informative observation is
that displacement and residual remain depth-structured even when rotational
scale is flat.
The middle-layer slowdown coincides with depths previously linked to more
language-neutral processing
\citep{wendler2024llamas,chang-etal-2022-geometry,bhattacharya-bojar-2023-unveiling},
and the elevated non-English residual is consistent with English-pivot
accounts~\citep{wendler2024llamas}.
We report these as associations.
Taken together, the results support a compact descriptive claim:
in coding and translation, residual-stream depth curves are model-dependent
and largely condition-stable, while content mainly rescales late-layer
amplitude.

\section{Related Work}
\subsection{Multilingual Representations}

Multilingual models exhibit structured geometry across depth: early and late
layers often retain language-specific structure, while middle layers form a
more shared, language-agnostic subspace
\citep{chang-etal-2022-geometry,bhattacharya-bojar-2023-unveiling}.
Large language models also frequently process non-English inputs by routing
through an English-centric latent space \citep{wendler2024llamas}.
Within that shared space, some computational features appear largely
independent of both the input language and the final output language
\citep{schut2025multilingual,kyriakou2026shared}.
These findings locate \emph{where} language-specific structure appears; they
do not measure how residual-stream geometry changes from one layer to the
next.

\subsection{Residual Streams, Readout, and Depth}

Mechanistic interpretability treats the residual stream as the transformer's
main communication channel, updated additively by attention and MLP blocks
\citep{elhage2021mathematical,olah2020overview}.
Layer-wise readout methods such as the logit lens and tuned lens ask what
each depth predicts about the next token
\citep{nostalgebraist2020logitlens,belrose2023eliciting}.
A complementary line of work debates the functional role of depth itself:
late layers have been argued to mainly refine output probabilities
\citep{csordas2026language}, while other results emphasise the need for
additional computational steps rather than additional parameters
\citep{saunshi2025reasoning}.
Related proposals allocate extra compute through looping or longer
generation
\citep{dehghani2018universal,wei2022chain}.
Our metrics address a different question: not what a layer predicts, but how
representations move between consecutive residual states.

\subsection{Mechanistic Interpretability and Reasoning Dynamics}

Mechanistic studies have reverse-engineered specific algorithmic operations
in transformers, such as Fourier multiplication circuits during grokking
\citep{nandaprogress} and the flow of operand information from early
attention into late MLP computations \citep{stolfo2023mechanistic}.
More recent work adopts geometric and kinematic views of reasoning.
Logical reasoning has been modelled as smooth geometric flows whose
structure shapes trajectory velocity and curvature
\citep{zhou2025geometry}.
\citet{fernando2025transformer} treat the residual stream as a dynamical
system with distinct phases of kinematic acceleration, while
\citet{damirchi2026truth} use discrete curvature to study model veracity,
arguing that scalar token kinematics alone can be confounded by lexical
effects.
These approaches motivate transition-centred analysis, but they primarily
track individual token paths rather than global cloud geometry across
depth.

\subsection{Geometric Analysis of Transformer Representations}

Other work studies the geometry of token representations more directly.
\citet{viswanathan2025geometry} characterise layer-wise token clouds using
intrinsic dimension and cosine similarity, relating spatial structure to
prediction loss; related efforts model cross-layer propagation with
dynamical systems \citep{shang2026unraveling}.
Directional consistency has also been measured across the sequence axis:
\citet{hosseini2026context} use cosine similarity of velocity vectors to
quantify trajectory straightening during in-context learning.
Much of this literature, however, treats each layer as a static snapshot.
Kinematic approaches track transitions, but usually for isolated tokens
rather than for the full token cloud.
We instead use orthogonal Procrustes analysis
\citep{schonemann1966generalized} to decompose each layer transition into
an optimal rigid rotation and a non-rigid residual.
This global, transition-centred view is what distinguishes our framework.

\subsection{Synthesis}

The multilingual literature maps where language-specific structure appears;
mechanistic and readout methods characterise what layers compute or predict;
geometric and kinematic studies describe representation shape or token
trajectories.
Fewer works ask how residual-stream geometry is organised across layer
transitions.
Our contribution is to measure that organisation directly: relative
displacement and Procrustes residual are strongly depth-structured, while
rotation magnitude is nearly constant across depth.
In short, prior work explains where language structure lives and what layers
do; we quantify how geometric change is allocated through the stack.

\section{Methods}
\subsection{Setup}

For each example, we record residual-stream activations for the input prompt
(\emph{prefill}) and for autoregressively generated tokens (\emph{generation}).
If a model has \(L\) layers and hidden dimension \(d\), token \(t\) is
represented by \(\mathbf{h}_t^\ell\in\mathbb{R}^d\) at depth \(\ell\).
Prefill yields one cloud of all prompt tokens after a single forward pass;
generation yields the hidden state of the newly produced token at each
decoding step.
Our analysis is transition-centered: instead of treating each layer as a
static snapshot, we measure how the token cloud is transformed from depth
\(\ell\) to \(\ell+1\).

\subsection{Geometric Metrics}

We view each layer transition as a geometric transformation of the residual
stream.
The metrics below separate how far tokens move, how sharply their trajectory
bends, how much of the cloud change is explained by a global rotation, and
how that rotation is distributed across planes.

\paragraph{Relative displacement.}
We measure each token's inter-layer movement relative to the norm of its source
representation:
\begin{equation}
    \delta_\ell(t)=
    \frac{\|\mathbf{h}_t^{\ell+1}-\mathbf{h}_t^\ell\|_2}
         {\|\mathbf{h}_t^\ell\|_2}.
    \label{eq:relative-displacement}
\end{equation}
This controls for growth in residual-stream norms with depth, which is common
in Pre-LN models: a large absolute step may still be small relative to a
growing residual vector.
Larger \(\delta_\ell\) means a larger relative update at that transition.
Unless noted otherwise, we report the mean of \(\delta_\ell(t)\) over tokens
within a prompt, then average across prompts and conditions.

\paragraph{Curvature.}
Treating depth as a trajectory, let
\(\mathbf{v}_\ell(t)=\mathbf{h}_t^{\ell+1}-\mathbf{h}_t^\ell\) be the update
at transition \(\ell\).
Curvature compares consecutive updates:
\begin{equation}
    \kappa_\ell(t)=
    \frac{
      \|(\mathbf{h}_t^{\ell+2}-\mathbf{h}_t^{\ell+1})
       -(\mathbf{h}_t^{\ell+1}-\mathbf{h}_t^\ell)\|_2
    }{
      \|\mathbf{h}_t^{\ell+1}-\mathbf{h}_t^\ell\|_2^2
    }.
    \label{eq:curvature}
\end{equation}
Large values indicate that the update changes strongly relative to its
magnitude, whether by turning direction or by changing step size.
We use curvature only as a descriptive complement to displacement.

\paragraph{Global rotation.}
At each transition, the tokens at depths \(\ell\) and \(\ell+1\) form two
point clouds in \(\mathbb{R}^d\).
We center both clouds and use orthogonal Procrustes alignment
\citep{schonemann1966generalized} to find the rotation \(R_\ell^*\) that best
maps the source cloud onto the target cloud.
We summarize the size of this rigid component by
\begin{equation}
    \mathcal{R}_\ell=\|R_\ell^*-I\|_F.
    \label{eq:rotation-magnitude}
\end{equation}
A larger value means a larger global reorientation of the token cloud.
Because \(R_\ell^*\) is orthogonal, \(\mathcal{R}_\ell\) depends only on how
far the fitted rotation lies from the identity, not on residual-stream scale.

\paragraph{Procrustes residual.}
A single global rotation does not explain every token's movement.
After applying \(R_\ell^*\), we compute the Euclidean distance between each
aligned source token and its centered target, then average across tokens
within each prompt and transition.
The residual is the part of the cloud change that cannot be absorbed by one
shared rotation: anisotropic stretching, local rearrangements, and other
non-rigid effects.
We treat it as a measure of non-rigid mismatch, not as a direct measure of
computational effort.

\paragraph{Rotation-angle concentration.}
A high-dimensional rotation may be spread across many planes or dominated by
a few.
Let \(D_\ell=R_\ell^*-I\) and \(q=\lfloor d/2\rfloor\).
We define
\begin{equation}
    \mathcal{C}_\ell=
    \frac{\|D_\ell\|_F^2}{q\|D_\ell\|_2^2}.
    \label{eq:angle-concentration}
\end{equation}
This is an efficient proxy derived from \(R_\ell^*-I\), not a full
eigendecomposition of the rotation.
Larger \(\mathcal{C}_\ell\) means rotational change is distributed across
more planes; smaller values mean fewer planes dominate.
Together with \(\mathcal{R}_\ell\) and the Procrustes residual, it lets us
ask whether constant rotation \emph{size} co-occurs with structured rotation
\emph{geometry}.

\subsection{Statistical Validation}
\label{sec:statistical-methods}

We analyze models and phases separately using the previously computed
prompt-level metrics; activations are not recomputed.
For each prompt, token-level values are averaged at every transition; prompts
are then averaged within condition, with conditions weighted equally.
This yields one mean depth curve per model, phase, and metric.

\paragraph{Depth modulation.}
For relative displacement, Procrustes residual, and angle concentration, we
score depth structure as the standard deviation of the mean curve divided by
its absolute mean.
A large score means some transitions are consistently larger than others.
We compare the observed score with \(2{,}000\) null scores obtained by
shuffling transition labels within each prompt, which preserves observed
values but removes shared depth order.
Holm-adjusted \(p<0.05\) indicates significant depth modulation.

\paragraph{Rotation constancy.}
For rotation magnitude, we ask whether the mean value across depth is
essentially constant.
We measure this with relative peak-to-trough variation:
\begin{equation}
    V=
    \frac{\max_{\ell}\bar{\mathcal{R}}_{\ell}
          -\min_{\ell}\bar{\mathcal{R}}_{\ell}}
         {|\operatorname{mean}_{\ell}(\bar{\mathcal{R}}_{\ell})|}.
\end{equation}
Before looking at the data, we defined rotation as practically constant when
\(V<0.03\) (less than \(3\%\) variation across depth).

We estimate uncertainty with \(2{,}000\) hierarchical-bootstrap samples,
resampling conditions and then prompts within conditions.
A model passes the constancy test only if both of the following hold after
Holm correction:
(i)~the one-sided 95\% upper bound on \(V\) is below \(0.03\), and
(ii)~the bootstrap probability that \(V\geq0.03\) yields an adjusted
\(p\)-value below \(0.05\).
The same bootstrap procedure provides confidence intervals for the other
reported effects.
Holm correction is applied separately within each metric family across
model--phase comparisons.

\paragraph{Dominant transitions and middle-layer slowdown.}
For relative displacement, Procrustes residual, and angle concentration, we
identify the transition with the largest mean value across depth and normalize
its index to \([0,1]\), where 0 is the first transition and 1 is the last.
This puts models with different depths on a common scale.
Bootstrap samples yield a 95\% interval for this normalized peak depth.

For relative displacement, we additionally test whether values are lower in
the middle third of depth than in the early and late thirds:
\begin{equation}
    S=
    \frac{
      \tfrac{1}{2}(\mu_{\mathrm{early}}+\mu_{\mathrm{late}})
      -\mu_{\mathrm{middle}}
    }{|\mu|},
    \label{eq:middle-slowdown}
\end{equation}
where thirds are defined on normalized depth and \(\mu\) is the mean across
all transitions.
The numerator compares the outer thirds with the middle; dividing by
\(|\mu|\) makes the score scale-free.
Positive \(S\) therefore indicates a middle-layer slowdown.
Coding and translation are tested separately in prefill and generation with
the same bootstrap and permutation procedures.
Holm correction is applied across model--task--phase comparisons.
Slowdown is supported when the adjusted \(p\)-value is below \(0.05\) and the
95\% bootstrap interval lies above zero.

\section{Experiments}

\subsection{Models}

Six instruction-tuned models across three families: Qwen3 (0.6B, 1.7B, 4B, 8B), Gemma-2-2B-IT, and StableLM-2-1.6B-Chat. All use Pre-LN architectures. This gives us variation in both family and parameter count\footnote{Implementation details \href{https://anonymous.4open.science/r/residual-stream-geometry-ED38/}{here}}.

\subsection{Tasks}

\paragraph{Code generation.} Problems from LiveCodeBench \citep{jain2025livecodebench}, stratified by difficulty (easy, medium, hard). Models generate Python solutions with greedy decoding up to 512 tokens. Monolingual (English in, English out).

\paragraph{Translation.} 100 sentence pairs per direction from Europarl v7 \citep{koehn-2005-europarl}: French, German, Czech, and Swedish paired with English (eight directions). Sentences under ten words filtered out. Greedy decoding up to 128 tokens. All four languages are Indo-European; we note this as a typological limitation.

\subsection{Activation Capture}

Forward hooks record the residual stream $H_\ell \in \mathbb{R}^{T \times d}$ after each layer's attention and MLP. For input: all $T$ prompt tokens from a single prefill pass. For generation: the hidden state of the last generated token at each autoregressive step. All runs use PyTorch MPS in \texttt{float16} on Apple M-series hardware. Layer indices are normalised to $[0, 1]$ for cross-model comparison.

\section{Results}
\subsection{Displacement}
\label{sec:displacement}

\begin{figure}[htbp]
  \centering
  \begin{subfigure}{0.48\textwidth}
    \centering
    \includegraphics[width=\linewidth]{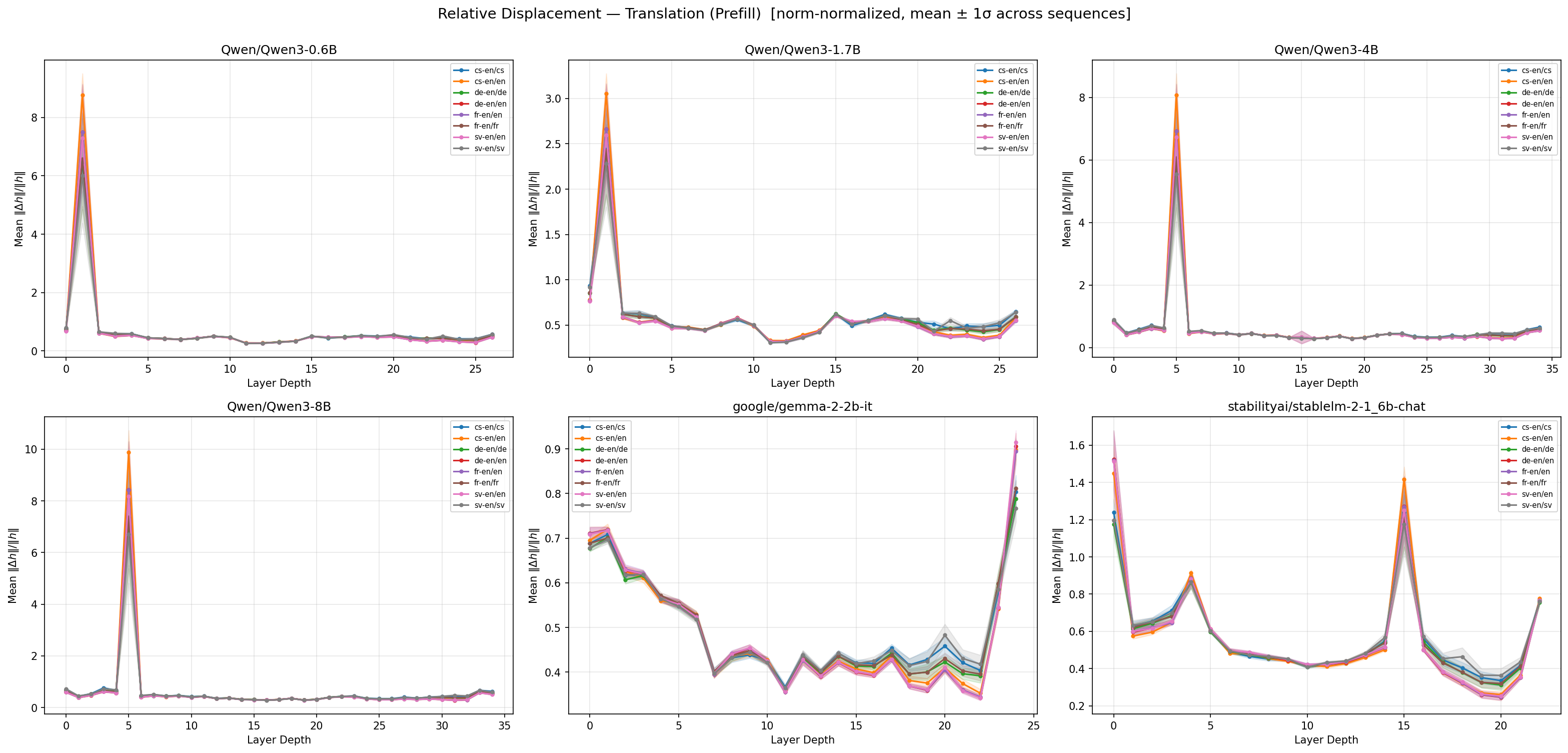}
    \caption{Translation}
    \label{fig:displacement-translation}
  \end{subfigure}\hfill
  \begin{subfigure}{0.48\textwidth}
    \centering
    \includegraphics[width=\linewidth]{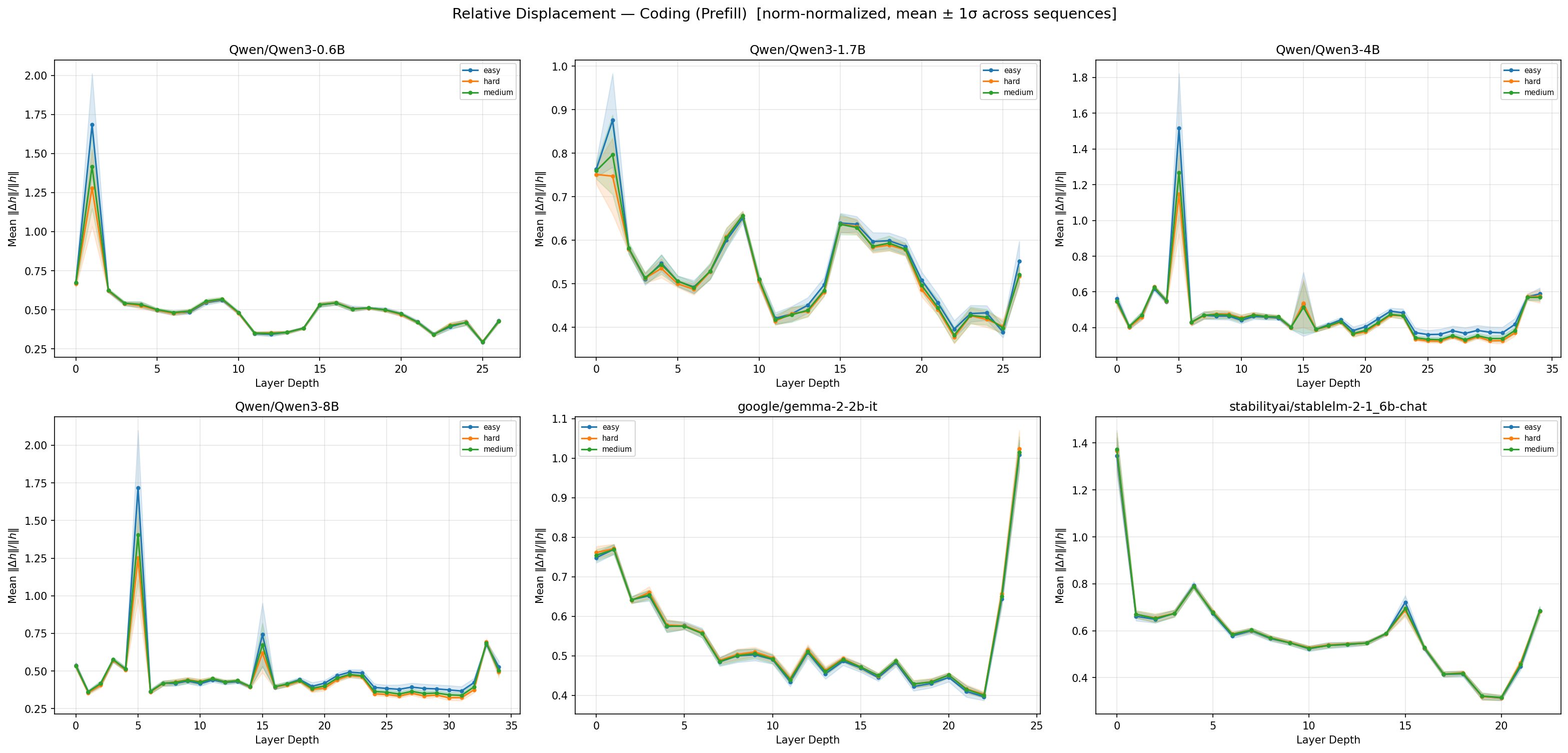}
    \caption{Coding}
    \label{fig:displacement-coding}
  \end{subfigure}
  \caption{Relative displacement during prefill. Within each model, the
    depth curve is nearly identical across conditions: the eight language
    directions (a) overlap, and coding difficulties (b) share the same
    shape, with only peak height varying.}
  \label{fig:relative-displacement}
\end{figure}

Relative displacement follows a model-specific depth curve that is largely
condition-invariant.
In prefill, the eight translation directions collapse onto one curve per
model (Figure~\ref{fig:displacement-translation}), and coding difficulties
reproduce the same shape (Figure~\ref{fig:displacement-coding}).
Qwen models are dominated by a sharp early peak; Gemma decays more smoothly
before a final rise; StableLM shows multiple early peaks and a middle valley.
Within each model, the \emph{shape} of change through depth is therefore more
stable across conditions than across models.

\begin{figure}[t]
  \centering
  \includegraphics[width=\linewidth]{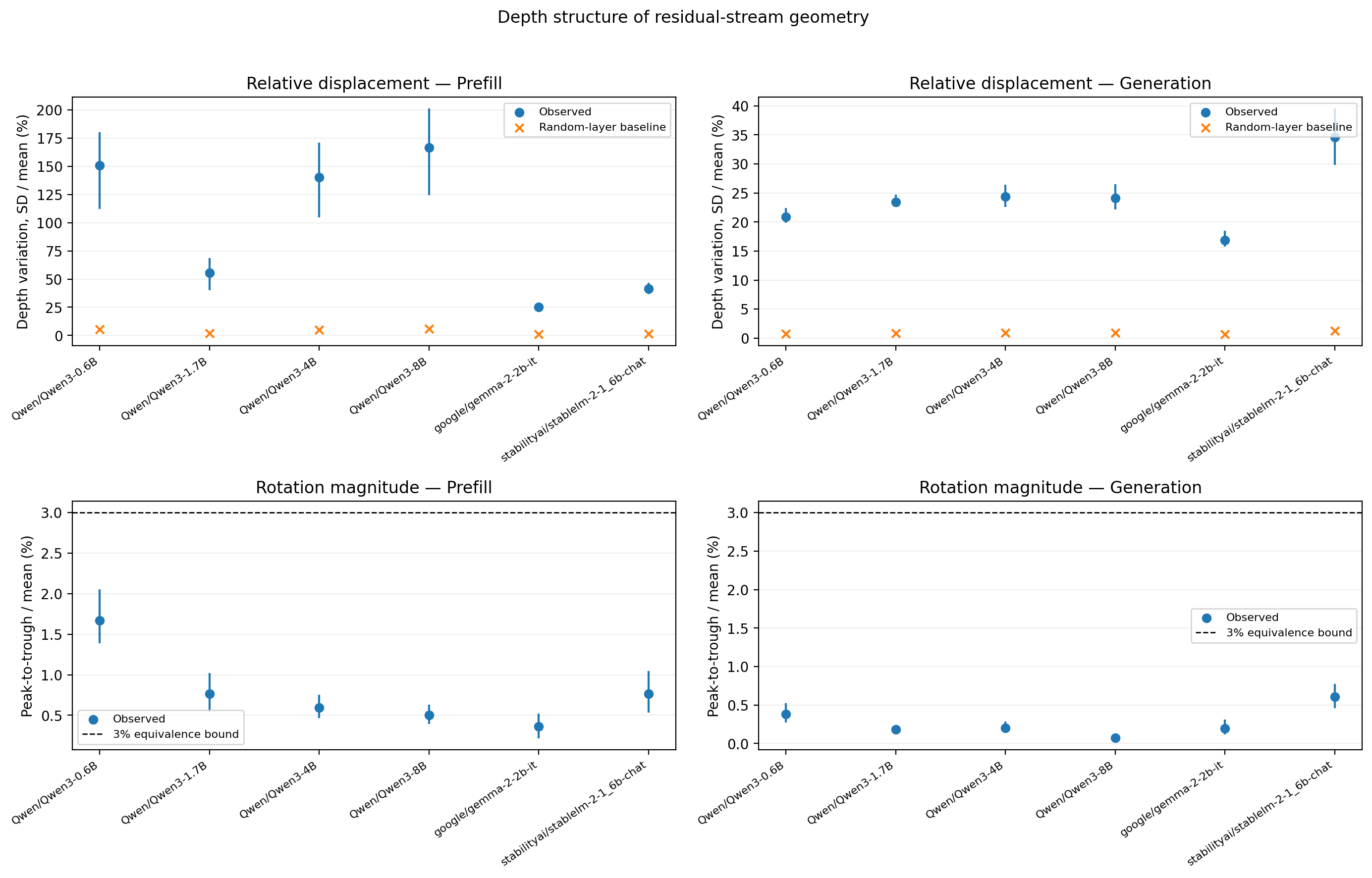}
  \caption{Depth structure of relative displacement (top) and rotation
    magnitude (bottom). Displacement is strongly modulated by depth;
    rotation magnitude stays below the predeclared \(3\%\) constancy bound
    in every model and phase.}
  \label{fig:depth-structure}
\end{figure}

This structure is statistically robust.
Depth modulation of relative displacement is significant for all six models
in both prefill and generation
(Holm-adjusted \(p<0.05\); Figure~\ref{fig:depth-structure}).
Prefill modulation is large (median \(97.8\%\); model range
\(25.2\%\)--\(166.4\%\)).
Depth structure persists in generation, though with smaller relative
variation across layers (median \(23.8\%\); \(16.9\%\)--\(34.6\%\)).

\begin{figure}[t]
  \centering
  \includegraphics[width=\linewidth]{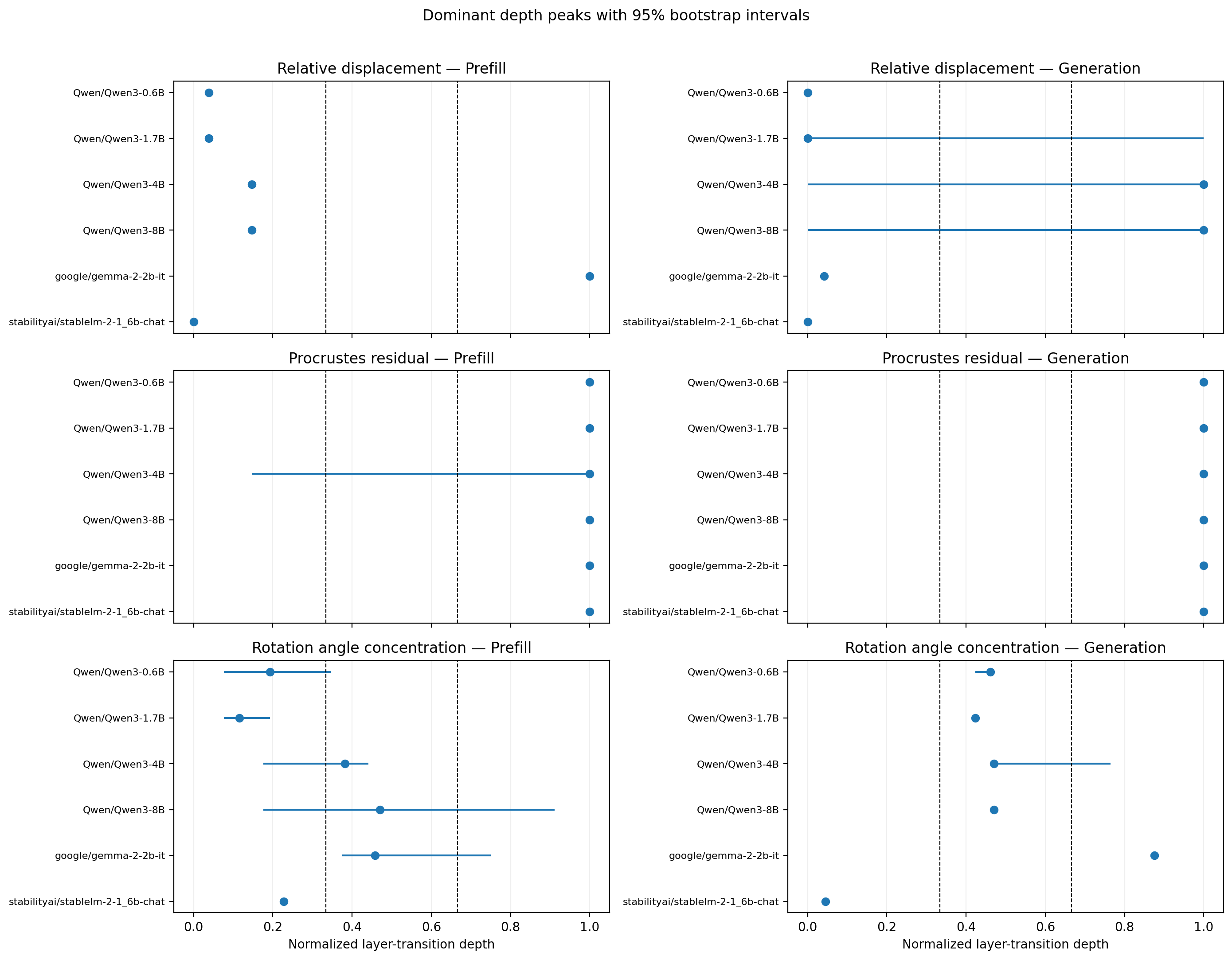}
  \caption{Normalized depth of the maximizing transition, with 95\%
    bootstrap intervals. Displacement peaks early for most models in
    prefill; Procrustes residual peaks at the final transition in every
    model and phase.}
  \label{fig:dominant-peaks}
\end{figure}

Dominant peaks during prefill fall in the early third for five of six models
(median normalized depth \(0.09\)); Gemma is the exception, peaking at the
final transition (Figure~\ref{fig:dominant-peaks}).
During generation, peak location is less stable for several Qwen models, with
bootstrap intervals that span much of the depth axis.

\begin{figure}[t]
  \centering
  \includegraphics[width=\linewidth]{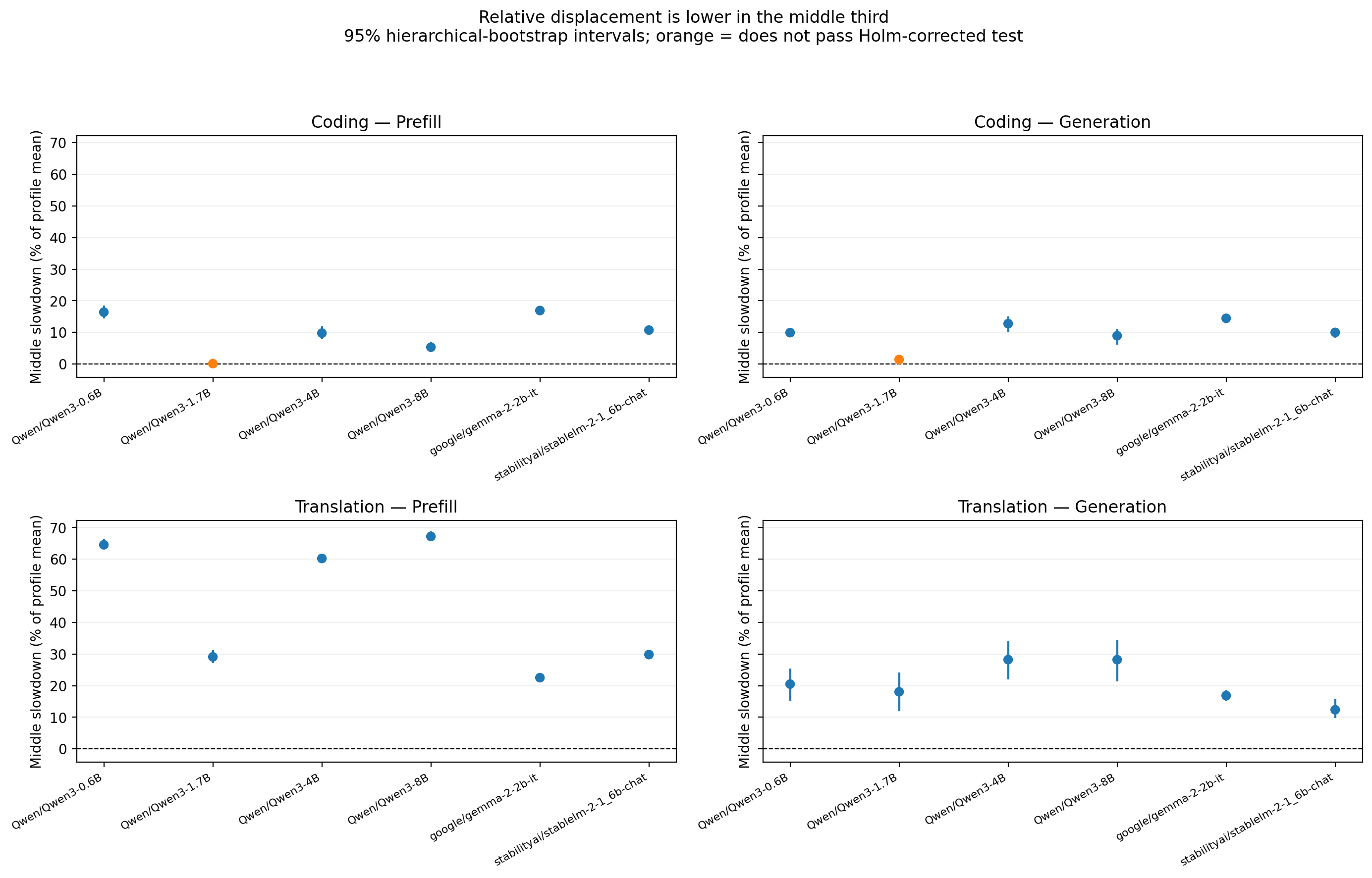}
  \caption{Middle-layer slowdown for relative displacement. Positive values
    indicate lower displacement in the middle third. Orange markers fail the
    Holm-corrected test.}
  \label{fig:middle-slowdown}
\end{figure}

A second regular feature is a middle-layer slowdown
(Figure~\ref{fig:middle-slowdown}).
Relative displacement is lower in the middle depth third than in the early
and late thirds for translation in all six models
(prefill median \(45.0\%\); generation \(19.3\%\)).
Coding shows the same sign but a smaller effect (medians \(\approx 10\%\));
five of six models pass the Holm-corrected test, with Qwen3-1.7B the sole
exception in both coding phases.
The early-middle-late rhythm is thus shared across the coding and
translation conditions studied here, while its strength depends on condition
and phase.

\begin{figure}[t]
  \centering
  \includegraphics[width=0.92\linewidth]{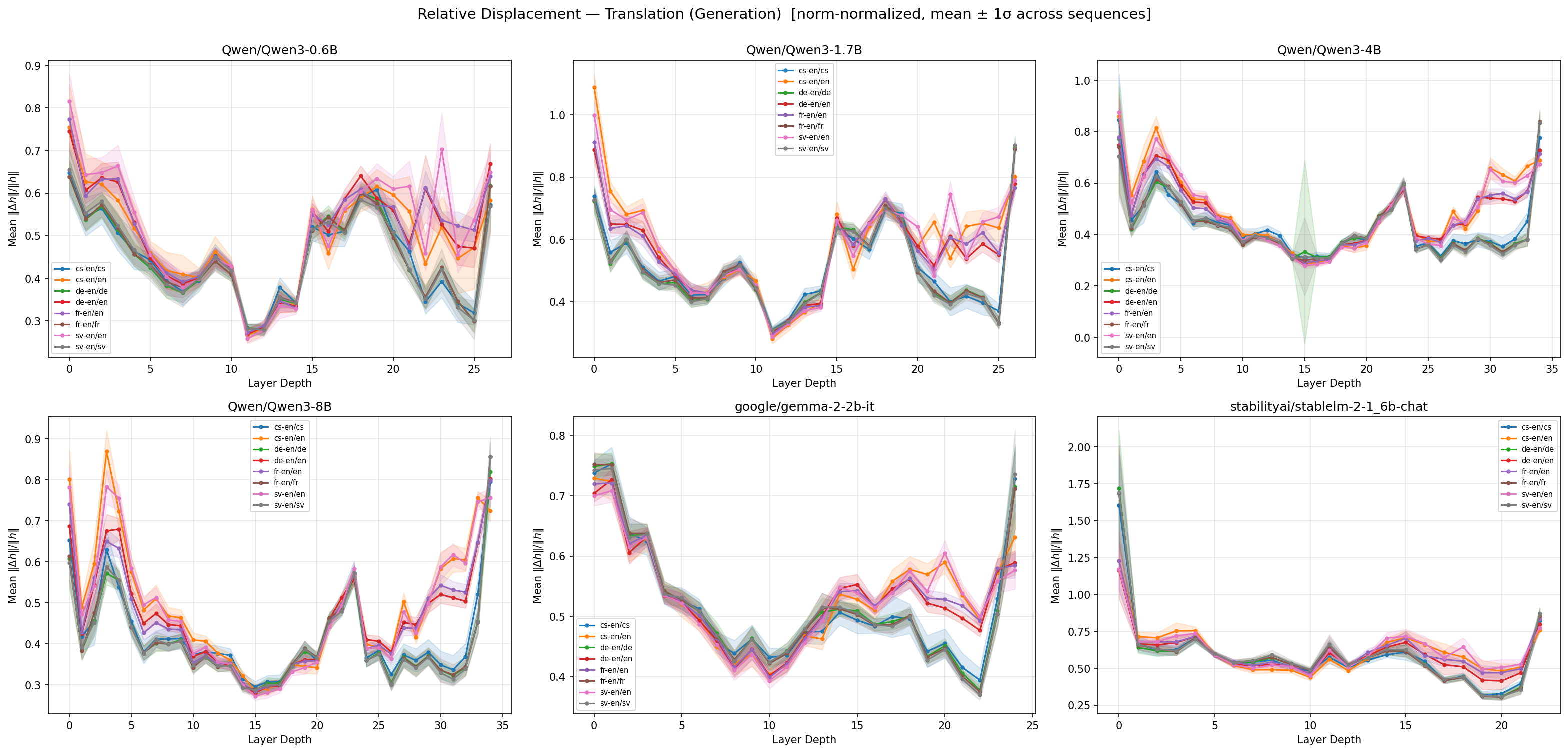}
  \caption{Relative displacement during translation generation.
    Early and middle layers remain nearly condition-invariant; final layers
    separate by target language, with larger updates for non-English targets.}
  \label{fig:displacement-generation}
\end{figure}

Generation makes the late-layer condition effect visible
(Figure~\ref{fig:displacement-generation}).
Early and middle layers still share one curve across directions, but
final-layer displacement splits by target language: non-English targets show
larger updates than English targets.
Coding generation preserves the same qualitative depth rhythm, with only
modest difficulty-related magnitude differences.
As shown next, the same English/non-English separation appears in Procrustes
residual.

\subsection{Rotation and Non-Rigid Structure}
\label{sec:rotation-results}

\begin{figure}[t]
  \centering
  \includegraphics[width=0.92\linewidth]{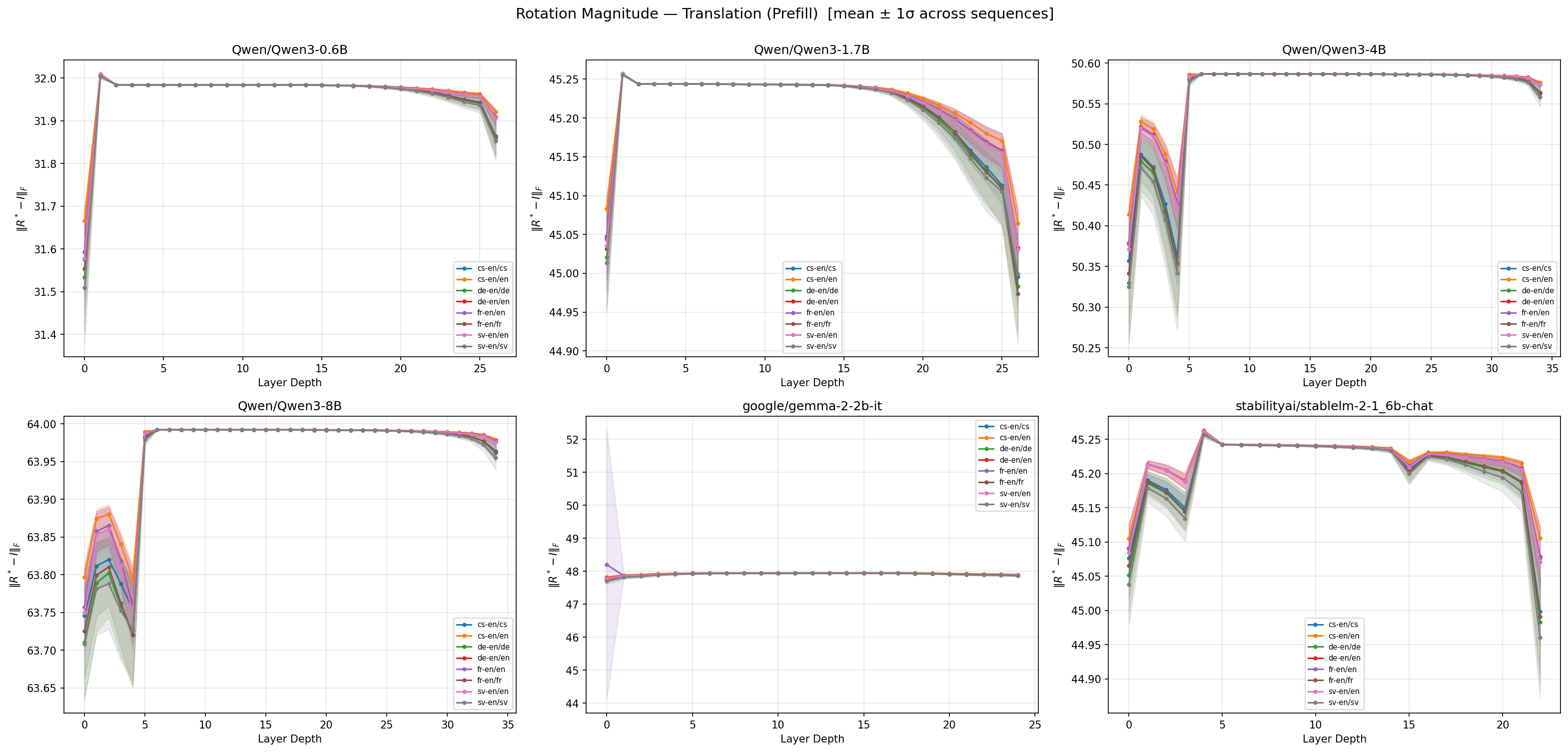}
  \caption{Rotation magnitude during translation prefill.
    Within each model the depth curve is nearly flat and overlaps across
    language directions; absolute scale grows with model size.}
  \label{fig:rotation-magnitude}
\end{figure}

\begin{table}[t]
\centering
\small
\setlength{\tabcolsep}{3pt}
\caption{Depth-structure tests (medians over models).
Modulation = SD/mean; rotation = peak-to-trough variation.
All comparisons pass after Holm correction
($\max p_{\mathrm{Holm}}=0.006$).}
\label{tab:depth-structure-summary}
\begin{tabular}{@{}l cc cc@{}}
\toprule
& \multicolumn{2}{c}{Prefill} & \multicolumn{2}{c}{Generation} \\
\cmidrule(lr){2-3}\cmidrule(lr){4-5}
Observable & Effect & Pass & Effect & Pass \\
\midrule
Rel.\ displacement
  & 97.8\% & 6/6
  & 23.8\% & 6/6 \\
Rotation magnitude
  & 0.68\% & 6/6
  & 0.20\% & 6/6 \\
Procrustes residual
  & 101.5\% & 6/6
  & 114.7\% & 6/6 \\
Angle concentration
  & 13.4\% & 6/6
  & 9.1\% & 6/6 \\
\bottomrule
\end{tabular}
\end{table}

Rotation magnitude is nearly constant across depth
(Figure~\ref{fig:rotation-magnitude}; Table~\ref{tab:depth-structure-summary}).
Peak-to-trough variation is below \(3\%\) for every model and phase
(prefill median \(0.68\%\); generation \(0.20\%\)), and all twelve
comparisons pass the practical-constancy test
(Figure~\ref{fig:depth-structure}, bottom).
Absolute magnitudes grow with hidden dimension \(d\), consistent with the
high-dimensional scale \(\|R-I\|_F\approx\sqrt{2d}\).
We therefore do not interpret flatness alone as evidence of a learned
computational invariant; in high dimensions, Procrustes alignments can yield
stable magnitudes for geometric reasons.
The depth curve itself, however, remains nearly flat across conditions.

\begin{figure}[t]
  \centering
  \includegraphics[width=\linewidth]{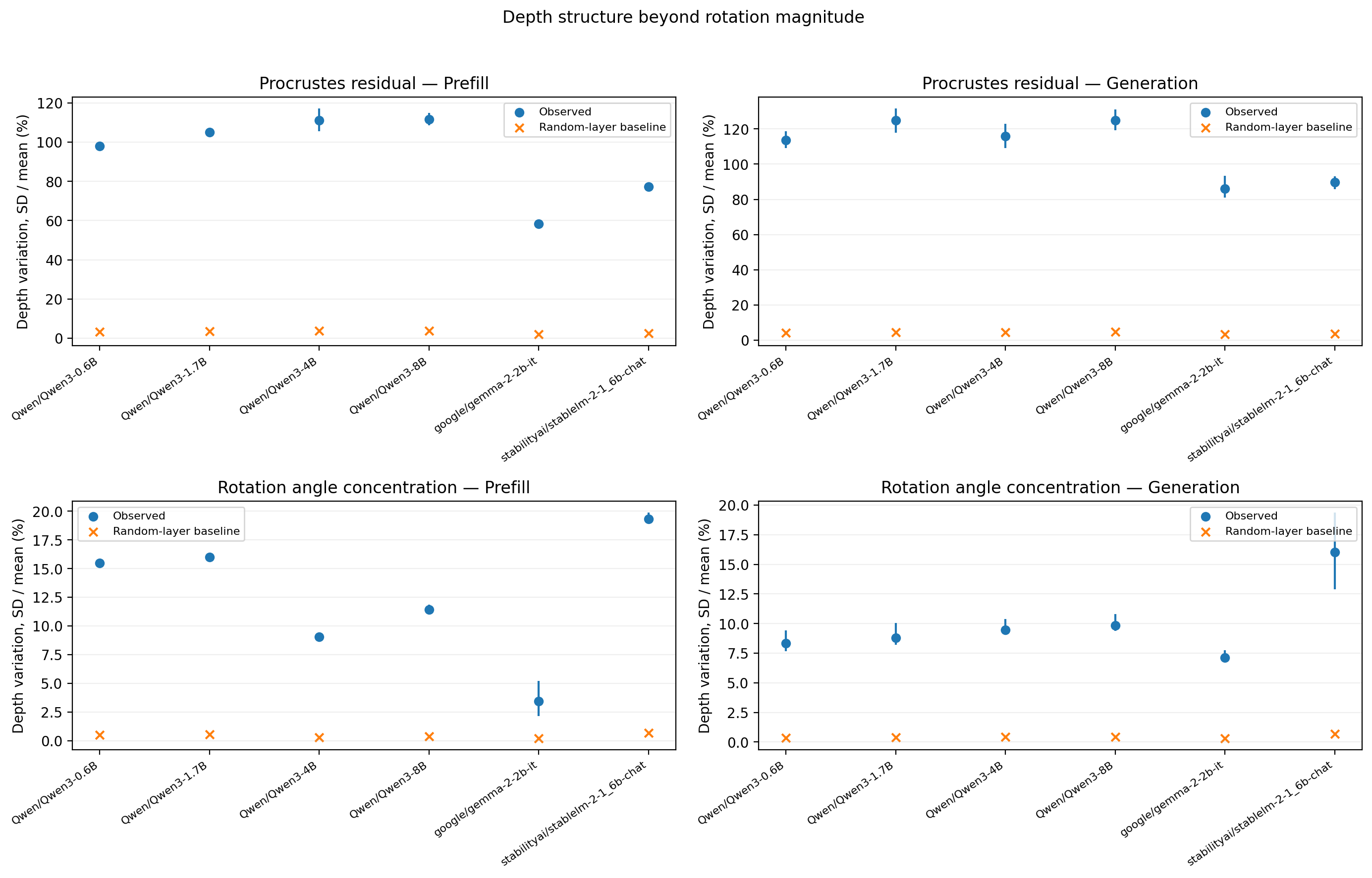}
  \caption{Depth structure beyond rotation magnitude. Procrustes residual
    and angle concentration are both significantly modulated by depth, even
    though rotation magnitude itself is nearly constant.}
  \label{fig:rotation-structure}
\end{figure}

\begin{figure}[t]
  \centering
  \includegraphics[width=0.92\linewidth]{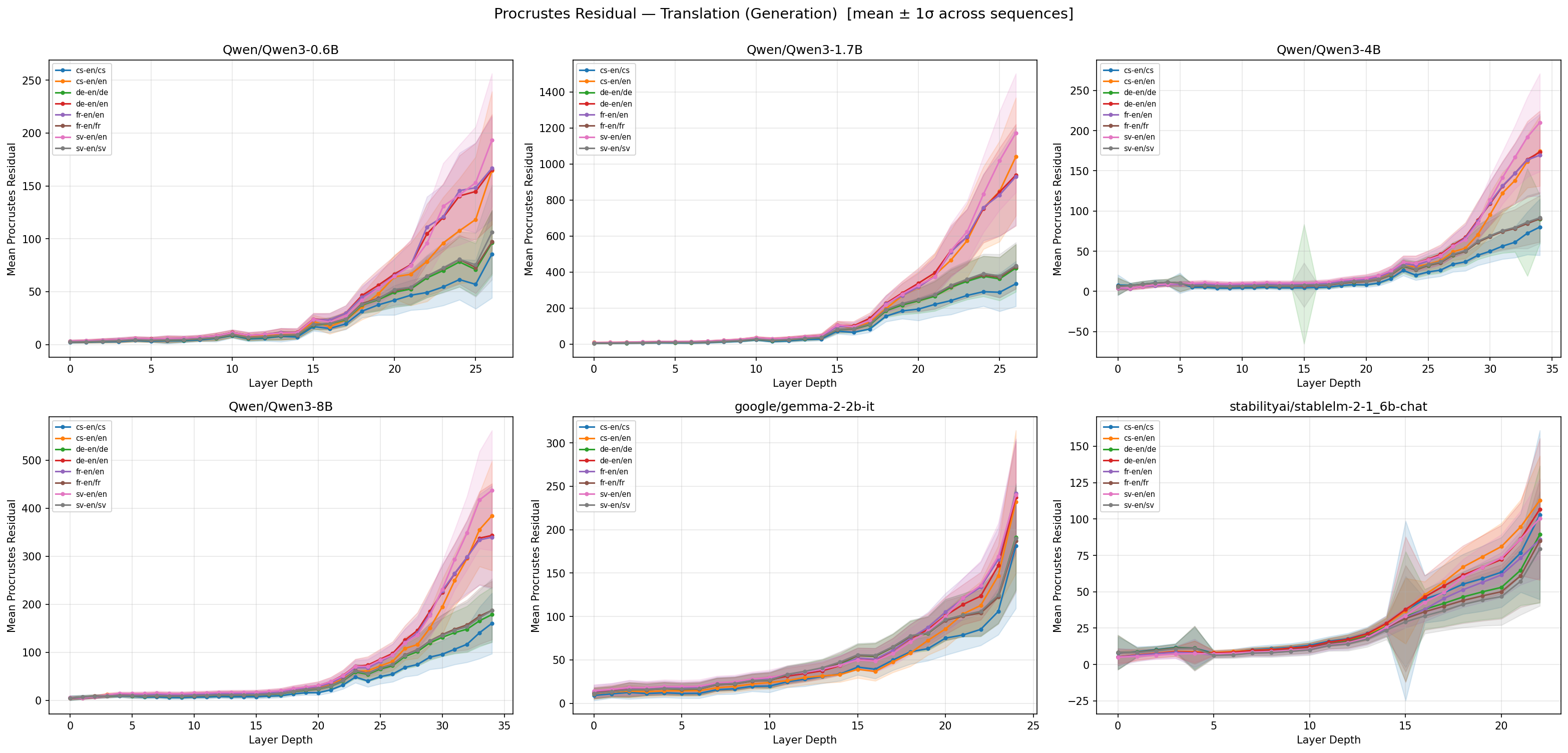}
  \caption{Procrustes residual during translation generation.
    Residual stays low through early and middle layers, then rises sharply
    near the output and separates by target language.}
  \label{fig:procrustes-generation}
\end{figure}

Near-constant magnitude does not imply unstructured geometry.
Procrustes residual is strongly depth-modulated in all models
(prefill median \(101.5\%\); generation \(114.7\%\);
Figure~\ref{fig:rotation-structure}).
Its maximizing transition is the final layer for every model in both phases
(normalized depth \(1.0\)).
During translation generation, residual curves stay low through most of the
stack and then rise sharply near the output
(Figure~\ref{fig:procrustes-generation}), again separating by target
language: non-English targets leave a larger non-rigid mismatch than English
targets.
As in Methods, we treat residual as geometric mismatch after global
alignment, not as computational effort.

\begin{figure}[t]
  \centering
  \includegraphics[width=0.92\linewidth]{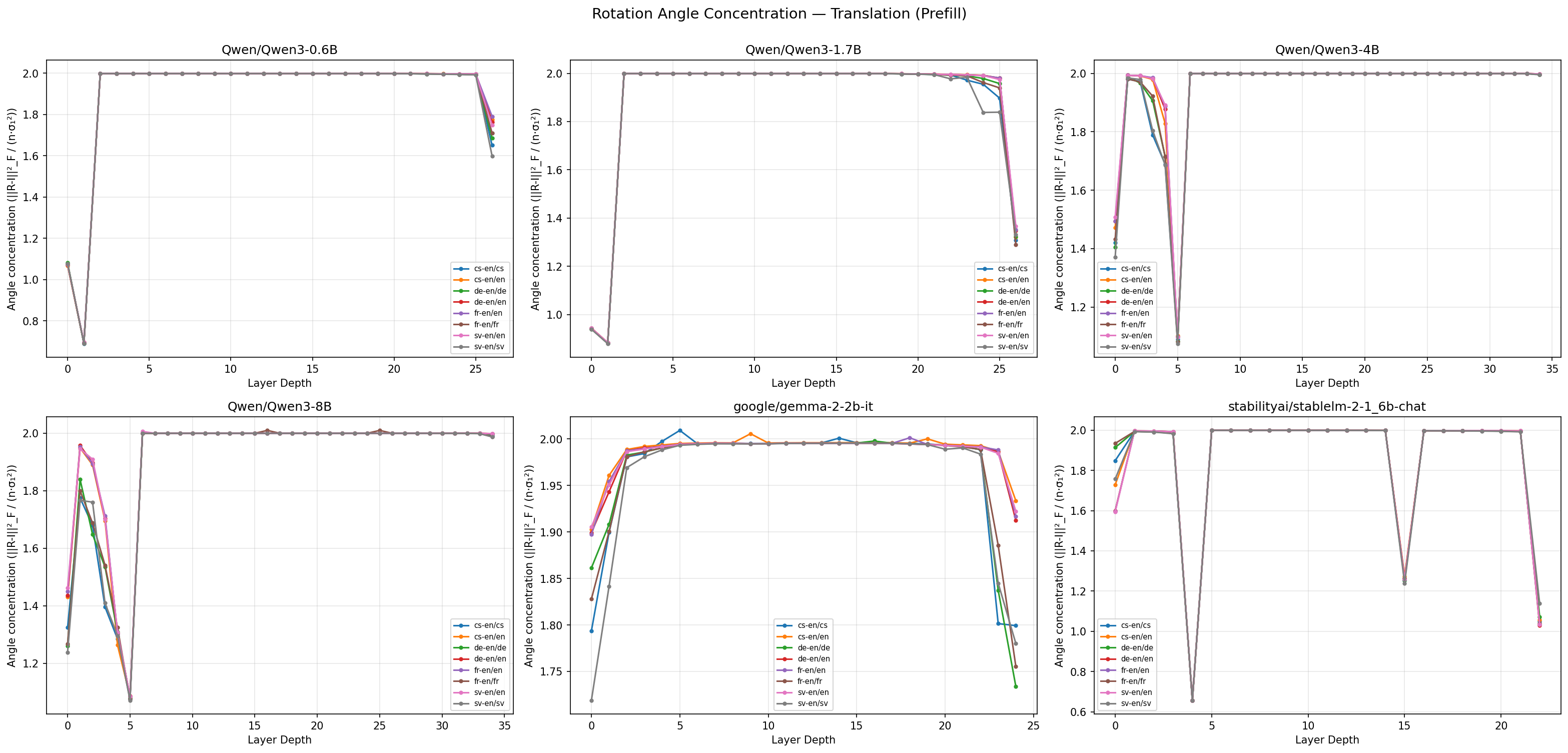}
  \caption{Rotation-angle concentration during translation prefill.
    Most models form a broad mid-depth plateau, with sharper changes in
    early and final transitions; language directions largely overlap.}
  \label{fig:angle-concentration}
\end{figure}

Angle concentration is also depth-structured, though more weakly
(Table~\ref{tab:depth-structure-summary}).
Prefill peaks fall early to mid-depth (median \(0.30\)); generation peaks
are more mid-depth (median \(0.47\)).
The raw curves show a broad mid-depth plateau with sharper changes near the
beginning and end of the stack
(Figure~\ref{fig:angle-concentration}).
Together these results separate two claims: the \emph{size} of the global
rotation is nearly depth-invariant, while the \emph{distribution} of that
rotation across planes and the residual mismatch after alignment both vary
systematically with depth.
The informative signal is this dissociation, not rotational flatness alone.

\subsection{Curvature}
\label{sec:curvature}

\begin{figure}[t]
  \centering
  \includegraphics[width=0.85\linewidth]{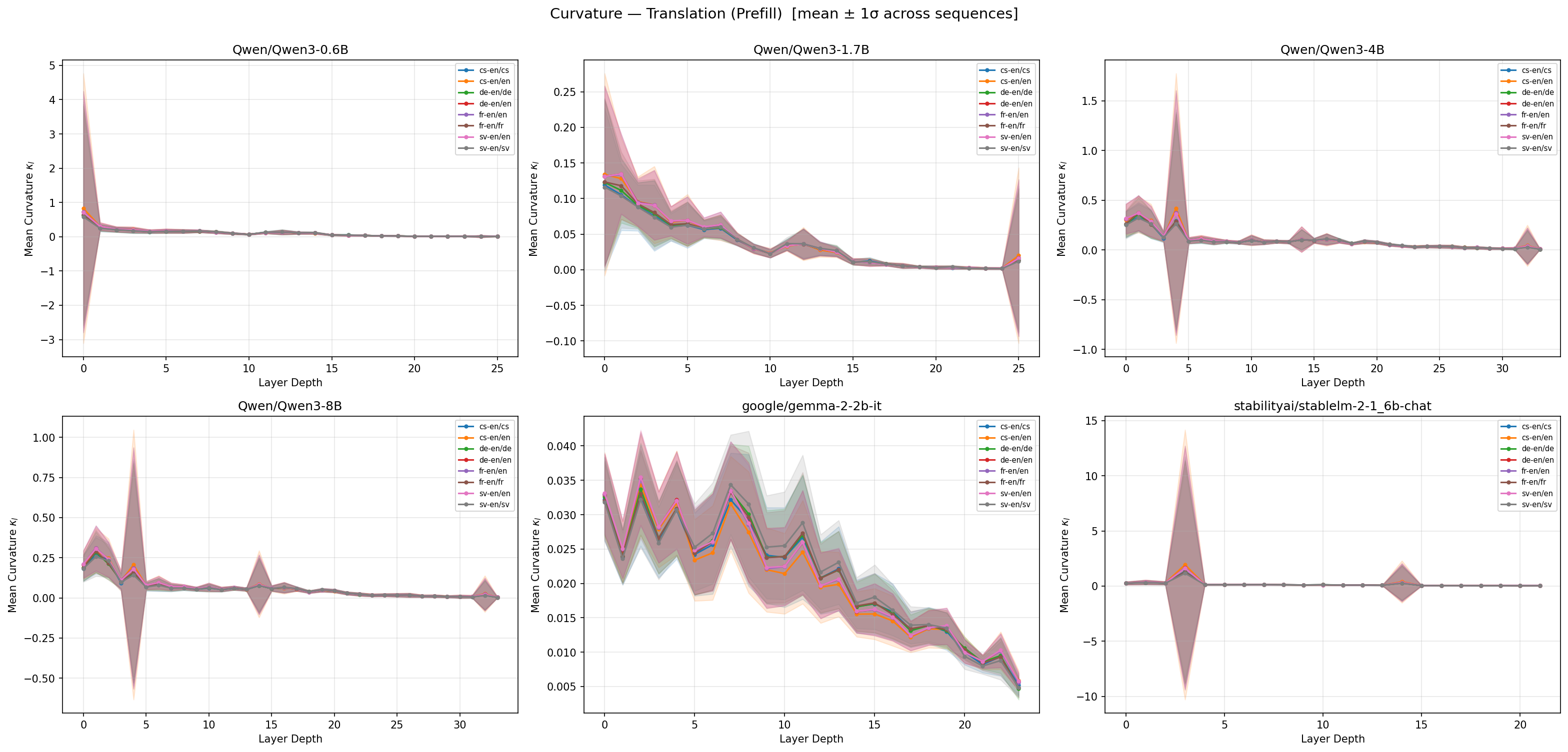}\\[2pt]
  \includegraphics[width=0.85\linewidth]{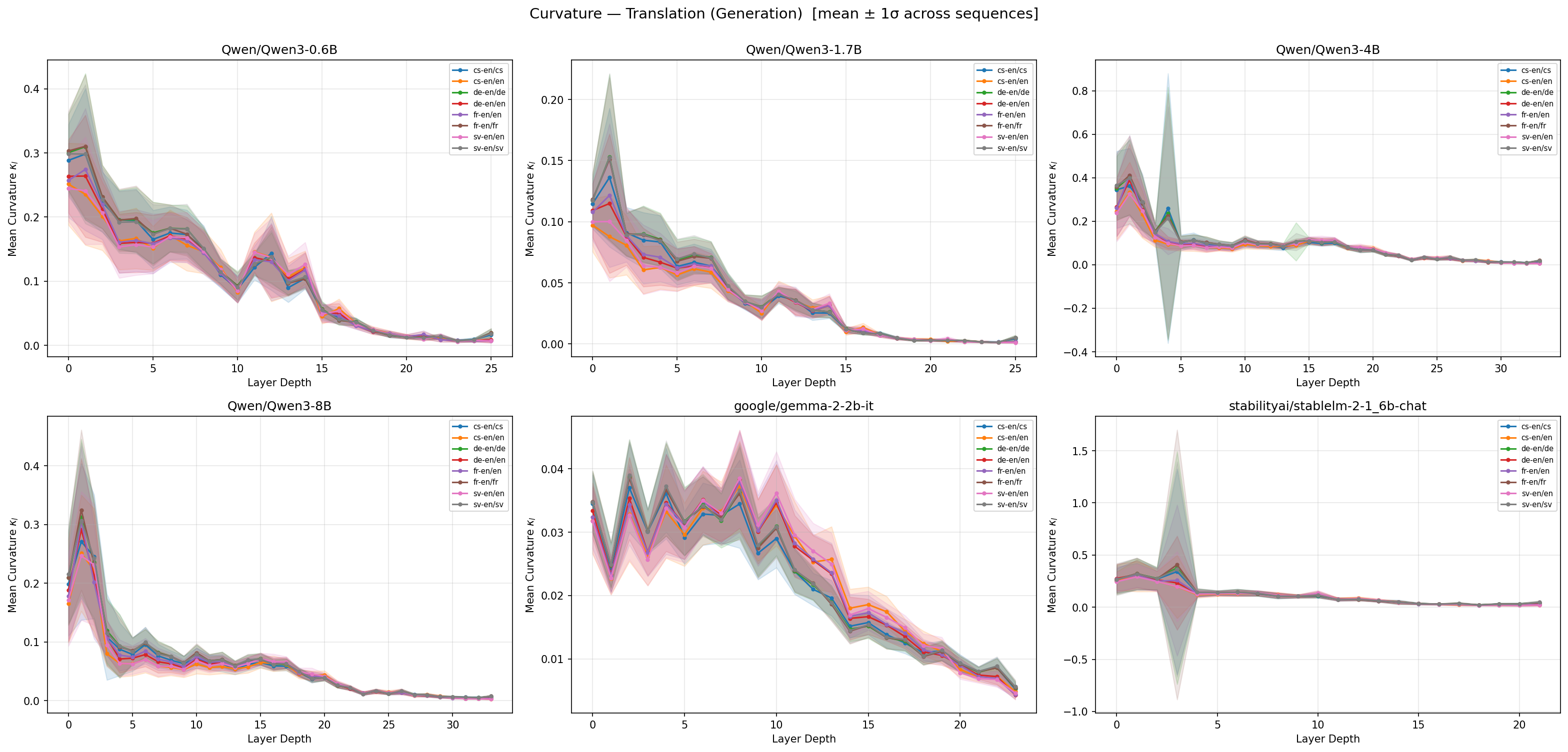}
  \caption{Curvature for translation (descriptive).
    Prefill is spike-dominated for Qwen/StableLM and smoother for Gemma;
    generation often shows an early-high, late-low decay.}
  \label{fig:curvature-prefill-generation}
\end{figure}

Curvature is used only descriptively and is not included in our statistical
tests.
Language directions again overlap within each model
(Figure~\ref{fig:curvature-prefill-generation}).
Prefill splits by model family: Qwen and StableLM show high-variance early
spikes, whereas Gemma decays more smoothly.
In generation, several models approach an early-high, late-low decay;
Qwen3-4B and StableLM remain spike-dominated.
Curvature is therefore a secondary complement to displacement, with more
model-to-model variability.

\paragraph{Summary.}
Treating residual streams as geometric transformations reveals nearly
constant rotational scale together with strongly depth-dependent
displacement and residual: early repositioning, a quieter middle third, and
rising late-layer non-rigid mismatch.
Across the coding and translation conditions studied here, depth curves are
model-dependent and largely condition-stable; conditions mainly rescale
late-layer amplitude rather than rewrite the depth schedule.

\section{Discussion}
\paragraph{Scope of claims.}
This paper offers a measurement framework and descriptive regularities, not a
mechanistic explanation.
The metrics quantify the geometry of residual-stream transitions; they are
not validated as measures of computational effort, reasoning difficulty, or
circuit activity.
Our evidence shows that, for code generation and translation, depth curves
are highly consistent across conditions within each model and differ across
models.
We therefore describe the schedule as \emph{model-dependent} and
\emph{condition-stable}, rather than as architecture-caused:
the experiments compare prompts within fixed models and do not manipulate
architecture.
Likewise, near-constant rotation magnitude should be interpreted cautiously.
In high dimensions, orthogonal Procrustes alignments often yield
$\|R-I\|_F$ values near the concentration scale $\sqrt{2d}$, so flatness may
partly be a geometric baseline rather than a learned computational invariant.
What is informative is the dissociation: even when rotational scale is nearly
constant, relative displacement, Procrustes residual, and angle concentration
remain strongly depth-structured.

\paragraph{What the measurements show.}
Under the transformation view, residual-stream change is not uniform across
depth.
Relative displacement is typically larger early and late, with a quieter
middle third; Procrustes residual rises sharply near the output; and rotation
magnitude stays nearly flat.
Within each model, coding difficulties and translation directions largely
share the same depth curve.
The clearest condition effect appears late in generation: non-English targets
show larger displacement and Procrustes residual than English targets.
This is consistent with English-centric accounts of multilingual processing
\citep{wendler2024llamas,conneau2020emerging,ahuja2023mega}, but it remains
an observational association restricted to the language pairs studied here.
These measurements complement residual-stream and readout work.
Mechanistic studies treat the residual stream as a communication channel
\citep{elhage2021mathematical,olah2020overview}; logit-lens methods ask what
each depth predicts
\citep{nostalgebraist2020logitlens,belrose2023eliciting}.
Our question is different: how representations move between layers.
The resulting early-middle-late pattern is compatible with late-layer
refinement of token probabilities \citep{csordas2026language} and with
arguments that some problems need additional compute steps
\citep{saunshi2025reasoning}, but compatibility is not explanation.

\paragraph{Hypotheses for future work.}
Several non-exclusive hypotheses are consistent with the measurements.
Early displacement peaks may reflect remapping from embeddings into a working
residual space.
Middle-layer slowdown may indicate gradual refinement with smaller relative
updates.
Late residual growth may reflect preparation for unembedding, where one
global rotation is insufficient.
The late English/non-English gap may reflect uneven output-space familiarity
rather than a different depth schedule.
Testing these hypotheses will require interventions that lie outside the present study.

\paragraph{Implications.}
Within the settings studied here, two practical points follow.
First, methods that allocate extra compute through longer generation or layer
looping
\citep{dehghani2018universal,wei2022chain} can be checked
against whether they preserve or disrupt this measured depth pattern.
Second, multilingual evaluation may benefit from layer-resolved geometric
observables: shared depth curves can coexist with larger late-layer residual
for non-English generation.
Neither point assumes that residual equals computational cost.

\section{Conclusion}
We introduced a geometric view of residual streams: each layer transition is
a transformation of the token cloud, measured by relative displacement and
decomposed by orthogonal Procrustes analysis into rigid rotation and
non-rigid residual.
Applied to six models on code generation and translation, this view reveals
reproducible depth regularities---early and late updates, a quieter middle
third, nearly constant rotation magnitude, and rising late-layer residual---
that are largely shared across conditions within each model.
Content mainly rescales late-layer amplitude in these settings, with
non-English generation showing larger final-layer displacement and residual
than English generation.
We present these findings as descriptive geometric measurements, not as
explanations of computation.
The contribution is a framework for quantifying residual-stream transitions
and evidence that, for the tasks studied here, depth curves are
model-dependent and largely condition-stable.

\section*{Limitations}
We evaluate only code generation and translation, with Indo-European
languages paired with English; broader tasks and language families remain
untested.

The metrics describe geometric change and are not validated as measures of
computational effort.
Near-constant rotation magnitude may partly follow from high-dimensional
Procrustes geometry, so our claim emphasises the dissociation with structured
displacement and residual rather than flatness alone.
Curvature is descriptive only.

All models are instruction-tuned Pre-LN decoders under greedy decoding, and
because we do not manipulate architecture, model-dependent schedules should
not be read as causal architectural effects.




\bibliography{custom}




\end{document}